\documentclass[conference]{IEEEtran}
\usepackage[square,sort,comma,numbers]{natbib}
\setcitestyle{square}

\usepackage{fancyhdr}

\def\BibTeX{{\rm B\kern-.05em{\sc i\kern-.025em b}\kern-.08em
    T\kern-.1667em\lower.7ex\hbox{E}\kern-.125emX}}
\IEEEoverridecommandlockouts
\usepackage{amsmath,amssymb,amsfonts}
\usepackage{algorithm,algpseudocode}
\usepackage{graphicx}
\usepackage{textcomp}
\usepackage{xcolor}
\usepackage{verbatim}
\usepackage{hyperref}
\usepackage{tikz,pgfplots}
\usetikzlibrary{patterns, spy}
\pgfplotsset{compat=1.9}
\usepackage{caption}
\usepackage{subcaption}
\usepackage{flushend}

\DeclareMathOperator*{\T}{T}
\DeclareMathOperator*{\Tinv}{T^{-1}}
\DeclareMathOperator*{\R}{R}
\DeclareMathOperator*{\F}{H}
\DeclareMathOperator*{\Proj}{P}
\newcommand{\frames}[3]{\underset{\text{#2} \leftarrow \text{#3}}{#1}}
\newcommand{\transform}[2]{\frames{\T}{#1}{#2}}
\newcommand{\transforminv}[2]{\frames{\Tinv}{#1}{#2}}
\newcommand{\rotation}[2]{\frames{\R}{#1}{#2}}
\newcommand{\coord}[1]{\underset{\text{#1}}{\F}}
\newcommand{\vecsub}[2]{\Vec{#1}_{\text{#2}}}
\newcommand{\vecnorm}[1]{\lVert#1\rVert}
\newcommand{\frsub}[2]{{#1}_\text{#2}}
\newcommand{\transpose}{^\top}

\newcommand{\subfigimg}[3][,]{%
  \setbox1=\hbox{\includegraphics[#1]{#3}}
  \leavevmode\rlap{\usebox1}
  \rlap{\hspace*{5pt}\raisebox{\dimexpr\ht1-1\baselineskip}{#2}}
  \phantom{\usebox1}
}

\usepackage{soul}

\fancypagestyle{FirstPage}{
\lfoot{\copyright 2023 IEEE.  Personal use of this material is permitted.  Permission from IEEE must be obtained for all other uses, in any current or future media, including reprinting/republishing this material for advertising or promotional purposes, creating new collective works, for resale or redistribution to servers or lists, or reuse of any copyrighted component of this work in other works.} 
}

\begin{document}

\title{Automated Multimodal Data Annotation via Calibration With Indoor Positioning System\\
}

\author{\IEEEauthorblockN{Ryan~Rubel, Andrew~Dudash, Mohammad~Goli, James~O'Hara, and Karl~Wunderlich} \IEEEauthorblockA{Autonomous Systems Research Center\\Noblis\\ Reston, Virginia\\Email: \{ryan.rubel,andrew.dudash,mohammad.goli,james.ohara,karl.wunderlich\}@noblis.org}}

\maketitle

\begin{abstract}

Learned object detection methods based on fusion of LiDAR and camera data require labeled training samples, but niche applications, such as warehouse robotics or automated infrastructure, require semantic classes not available in large existing datasets. Therefore, to facilitate the rapid creation of multimodal object detection datasets and alleviate the burden of human labeling, we propose a novel automated annotation pipeline. Our method uses an indoor positioning system (IPS) to produce accurate detection labels for both point clouds and images and eliminates manual annotation entirely. In an experiment, the system annotates objects of interest 261.8 times faster than a human baseline and speeds up end-to-end dataset creation by 61.5\%.


\end{abstract}


\section{Introduction}
\thispagestyle{FirstPage}

The complementary characteristics of LiDARs and cameras motivate their combined use in object detection, especially for autonomous vehicles. Both modalities are complex and high-dimensional, so detectors rely on learned methods and, by extension, annotated datasets for their training and evaluation. However, collecting a multimodal object detection dataset requires careful sensor calibration and synchronization as well as ground truth labels for each modality. Manual annotation in both sensor frames is laborious, time-consuming, and expensive. There exist methods to partially or completely automate the process, yet most require either partial manual annotation or rely on an existing trained detector, which may not exist for niche objects.

To generalize to new objects and eliminate manual annotation altogether, we propose a method that instead relies on real-time position estimates from an indoor positioning system (IPS). The IPS requires setup prior to use and limits data collection to controlled environments, but our method is well-suited to a laboratory setting. To our knowledge, our system is the first pipeline capable of automatically estimating object poses and bounding boxes in an indoor or GPS-denied environment.

In this paper, we first explore existing automated annotation methods and explain the benefits of our own. Then, we describe the network of transformations that enables our method and detail our calibration techniques. We introduce a planar constraint to reduce the impact of sensor noise when calibrating with respect to the IPS. We outline the RANSAC algorithm we adapt to refine our 3D annotations and justify our choice of fitness function. Finally, we quantitatively and qualitatively analyze the speed and accuracy of the automated annotation system.

\begin{figure}
\centering
\begin{tabular}{@{}p{0.45\linewidth}@{\quad}p{0.45\linewidth}@{}}
\includegraphics[width=0.225\textwidth]{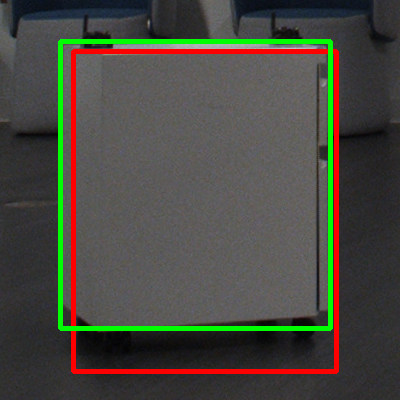} &
\includegraphics[width=0.225\textwidth]{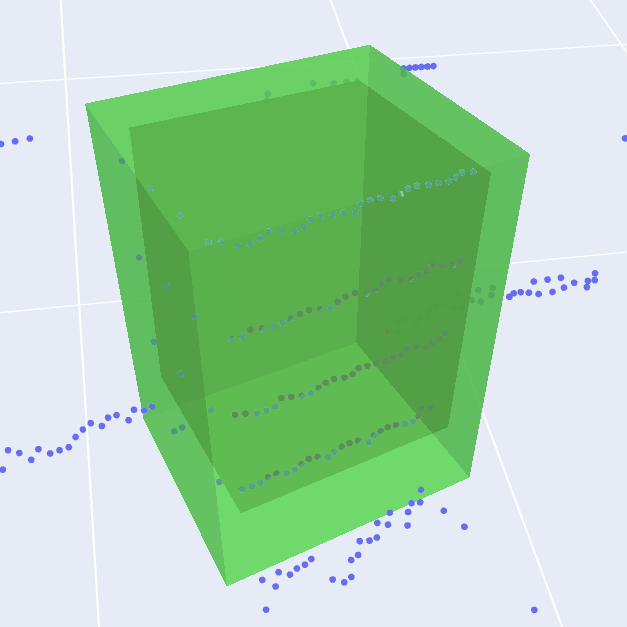} \\
\end{tabular}
\caption{Sample manual and automated annotations (best viewed in color). The green labels were manually annotated; the red labels were generated automatically.}
\end{figure}

\section{Related Work}

Due to safety and privacy concerns as well as practicality, ground truth generation in the wild is typically unable to rely on positioning sensors. For this reason, most existing automated annotation approaches rely on pretrained detectors or some degree of human effort. For instance, Marion \cite{8460950} automatically refines initial pose estimates provided by humans. Since they scan scenes with an RGBD camera, their point clouds are dense enough for  pose refinement through ICP \cite{CHEN1992145} \cite{121791}. We obtain sparser point clouds from a 16-channel LiDAR, so we instead opt for the more robust RANSAC \cite{10.1145/358669.358692}. Ince \cite{Ince_2021_ICCV} manually labels single video frames and uses a tracking algorithm to propagate those labels to subsequent frames. Watanabe \cite{WATANABE20201763} filters datamined images based on cosine similarity with manually selected exemplar samples and uses Mask R-CNN \cite{8237584} to automatically annotate unseen images. Hajri \cite{hajri2018automatic} uses high-precision positioning systems placed in cars to generate ground truth annotations. However, the quality of their annotations relies solely on the accuracy of these positioning sensors; their method has no refinement stage to ensure consistent labels.

Unlike Marion, Ince, or Watanabe, our approach requires no human annotation or existing detectors. Further, unlike Hajri, our system works in an indoor environment and is applicable for any object of interest sufficiently larger than a positioning sensor (hereafter referred to as an ``IPS beacon'').

\section{Method}

We automatically label the training data in two phases: collection and postprocessing. In the collection phase, we estimate the positions of our sensors and objects of interest with IPS beacons (Section \ref{section:frame_construction}), calibrate the different sensor frames based on their relative locations (Section \ref{section:camera_ips_calibration}), and then simultaneously collect data from the sensors and IPS beacons. In the postprocessing phase, we increase the accuracy of label bounding boxes by shifting the boxes to overlap with more LiDAR points (Section \ref{section:ransac_lidar_refinements}). The box shifting is done with RANSAC, an optimization algorithm that fits a model to data by fitting many randomly chosen subsets of the data and selecting the model that maximizes a fitness function.

\begin{figure}
\centering
\includegraphics[width=\linewidth]{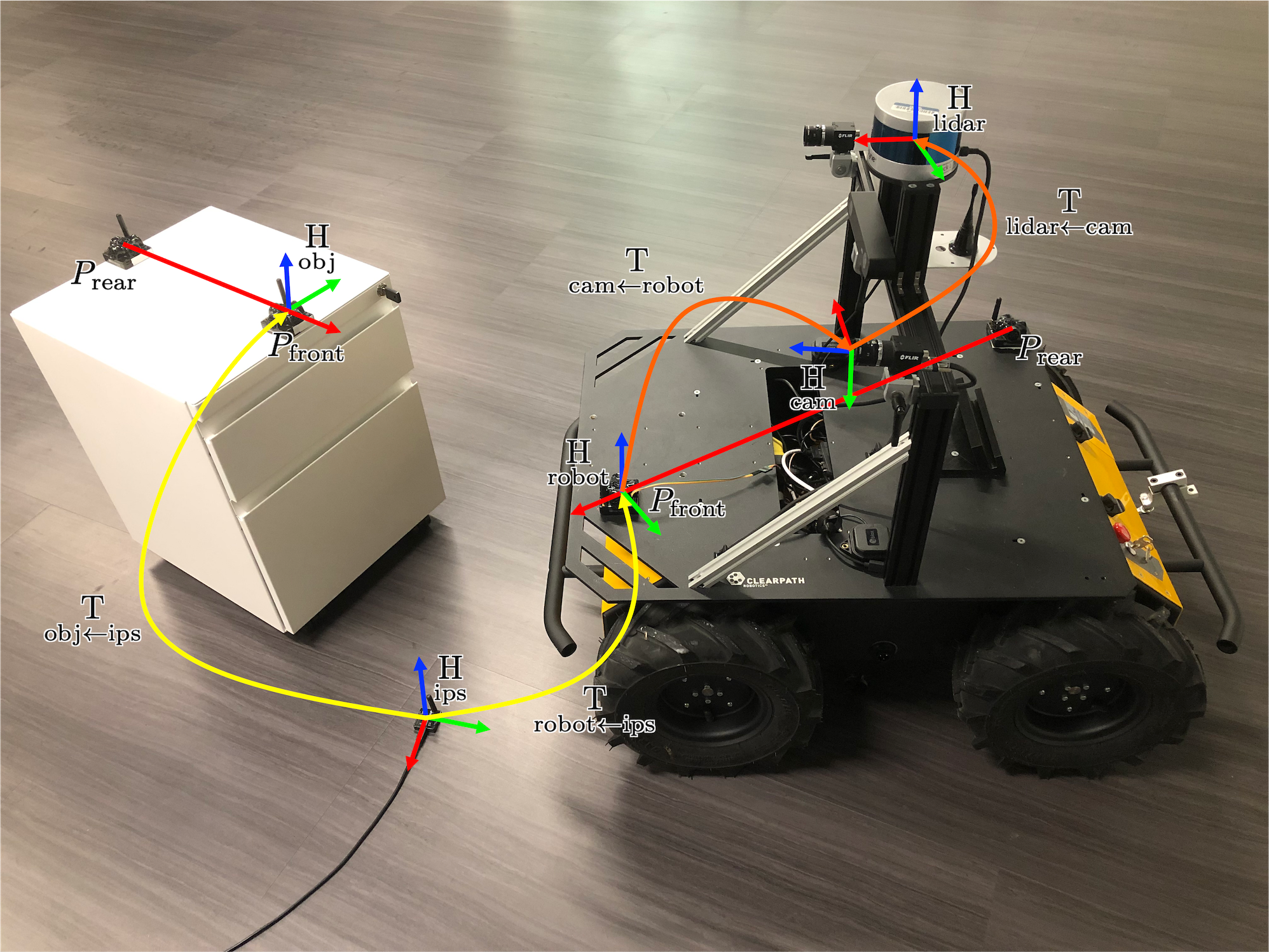}
\caption{Transformation graph superimposed over robot and object of interest. Orange transformations are calibrated offline. Yellow transformations are estimated online from IPS measurements, allowing real-time pose estimation of the object of interest (the cabinet shown here).}
\label{calib_network}
\end{figure}

\subsection{Network of Calibrated Transformations}
We automatically label data through a network of calibrated and measured transformations (see Fig. \ref{calib_network}). Although a conventional IPS beacon only estimates its position, we are able to derive its 4-DOF ($x$, $y$, $z$, yaw) pose by pairing it with another beacon (Section \ref{section:frame_construction}). We place one beacon frame $\coord{robot}$ on the base of the robot and an additional beacon frame $\coord{obj}$ on each object of interest. The poses of $\coord{robot}$ and each $\coord{obj}$ are estimated online by the indoor positioning system, enabling real-time estimation\footnotemark of the object's robot-centric pose as:

\begin{equation}
\transform{obj}{robot} = \transform{obj}{ips} \cdot \transforminv{robot}{ips}
\end{equation}

\footnotetext{$\transform{cam}{ips}$ could instead be calibrated directly, removing the intermediate frame $\coord{robot}$. This would bypass $\transform{robot}{ips}$ (estimated online from noisy IPS measurements) and thereby reduce the overall error of the setup. However, such a modification would require that $\transform{cam}{ips}$ be recalibrated with every movement of the robot, making it impractical.}

For use in autonomous perception, the generated ground truth pose of the object of interest must be transformed into both the LiDAR and camera frames--we require $\transform{obj}{cam}$ and $\transform{obj}{lidar}$. Since both the camera and $\coord{robot}$ are rigidly attached to the robot base, the camera-centric pose of the object of interest may be computed as:

\begin{equation}
\transform{obj}{cam} = \transform{obj}{robot} \cdot \transform{robot}{cam}
\end{equation}

The object's position on the image plane may then be found by applying the camera's projection matrix to the position component of the object's pose:

\begin{equation}
p_{\text{obj}} = \Proj \cdot \transform{obj}{cam} \cdot \begin{bmatrix}0 & 0 & 0 & 1 \end{bmatrix}\transpose
\end{equation}

Since the LiDAR is rigidly attached to the robot base, the object's LiDAR-centric pose may be estimated\footnotemark as:

\begin{equation}
\transform{obj}{lidar} = \transform{obj}{cam} \cdot \transform{cam}{lidar}
\end{equation}

\footnotetext{In principle, estimating $\transform{obj}{lidar}$ as $\transform{obj}{robot} \cdot \transform{robot}{lidar}$ would be more direct and therefore more precise. However, doing so would require either directly calibrating the LiDAR with IPS beacons (impractical due to the beacons' small size and the sparsity of observed point clouds) or jointly calibrating the LiDAR-camera-IPS setup, which is feasible, but complex. Instead, we rely on an existing LiDAR-camera calibration procedure, which is far simpler.}

We calibrate the unknown constants $\Proj$, $\transform{lidar}{cam}$, and $\transform{robot}{cam}$ offline. For the projection matrix $\Proj$, we use the OpenCV implementation of the standard checkerboard method \cite{camera_calibration}. Then, we follow \cite{lidar_camera_calibration} to find $\transform{lidar}{cam}$. Finally, we estimate $\transform{cam}{robot}$ as described in Section \ref{section:camera_ips_calibration}.

\subsection{Frame Construction from IPS Beacon Positions}
\label{section:frame_construction}
A conventional IPS beacon only estimates position, rather than pose. However, we may construct a pose by pairing two beacons\footnotemark, $\frsub{P}{front}$ and $\frsub{P}{rear}$. To derive a frame $\coord{frame}$ from two points, we constrain its $z$ axis to be parallel to that of the global IPS frame. Therefore, we replace the $z$ coordinates of both $\frsub{P}{front}$ and $\frsub{P}{rear}$ with their average:

\footnotetext{If we used three beacons for each frame, we could estimate full 6-DOF poses. However, for many autonomous perception tasks, roll and pitch are ignored.}

\begin{equation}
z_{\text{front}}, z_{\text{rear}} \gets
\frac{1}{2}(z_{\text{front}} + z_{\text{rear}})
\end{equation}

To derive $\rotation{frame}{ips}$, the rotation from the IPS frame to the beacon frame, we require three mutually orthogonal unit vectors describing the axes of $\coord{frame}$. We take the direction pointing from the rear beacon to the front beacon as the $x$ axis:

\begin{equation}
\hat{x} = \frac{\frsub{P}{front} - \frsub{P}{rear}}{\vecnorm{\frsub{P}{front} - \frsub{P}{rear}}}
\label{eq_xvec}
\end{equation}

We assume that the $xy$ plane of $\coord{frame}$ is parallel to the $xy$ plane of the global IPS frame. This means that the $z$ axes of these two frames must be parallel. Therefore, by construction, we have:

\begin{equation}
\hat{z} =
\begin{bmatrix}0 & 0 & 1 \end{bmatrix}\transpose
\end{equation}

Since the $y$ axis must be mutually orthogonal with both the $x$ and $z$ axes, we construct it via the cross product:

\begin{equation}
\hat{y} = \hat{x} \times \hat{z}
\end{equation}

We now assemble the desired rotation matrix:

\begin{equation}
\rotation{frame}{ips} =
\begin{bmatrix}
\uparrow & \uparrow & \uparrow\\
\hat{x} & \hat{y} & \hat{z}\\
\downarrow & \downarrow & \downarrow
\end{bmatrix}
\end{equation}

The front beacon then becomes the origin of the frame:

\begin{equation}
\transform{frame}{ips} =
\begin{bmatrix}
\rotation{frame}{ips} & \frsub{P}{front}\\
\Vec{0} & 1
\end{bmatrix}
\end{equation}

We position beacons within each frame so as to maximize their pairwise distance. This improves pose estimation accuracy, as is made clear with the introduction of explicit noise terms to \eqref{eq_xvec}:
\begin{equation*}
\hat{x} = \frac{(\frsub{P}{front} + \vecsub{\nu}{front}) -
(\frsub{P}{rear} + \vecsub{\nu}{rear})}{\vecnorm{(\frsub{P}{front} + \vecsub{\nu}{front}) - (\frsub{P}{rear} + \vecsub{\nu}{rear})}} =
\frac{\frsub{P}{front} - \frsub{P}{rear} + \Vec{\nu}}{\vecnorm{\frsub{P}{front} - \frsub{P}{rear} + \Vec{\nu}}}
\end{equation*}

Since $\vecsub{\nu}{front}$ and $\vecsub{\nu}{rear}$ do not (in general) depend on $\frsub{P}{front}$ and $\frsub{P}{rear}$, the true direction vector $\frsub{P}{front} - \frsub{P}{rear}$ comes to dominate the expression as $\vecnorm{\frsub{P}{front} - \frsub{P}{rear}}$ grows.

\subsection{Camera-IPS Calibration}
\label{section:camera_ips_calibration}
We now estimate $\transform{cam}{robot}$, the transformation from the frame defined by the two beacons mounted on the robot's base to the camera. Estimating the camera's pose in terms of IPS beacons allows the robot to be repositioned during data collection without requiring recalibration, as the robot's new pose may be estimated from the IPS.

The image projection of an IPS beacon follows the equation:
\begin{equation}
    p_{\text{beacon}} = \Proj \cdot \transform{cam}{robot} \cdot \transform{robot}{ips} \cdot P_{\text{beacon}}
\label{eq_beacon_proj}
\end{equation}

where $P_{\text{beacon}}$ denotes the beacon's position in the IPS frame. $\transform{robot}{ips}$ may be computed from the IPS beacons mounted on the robot as described in Section \ref{section:frame_construction}. To reduce error propagation from noisy $\transform{robot}{ips}$ estimates, we keep the robot stationary throughout the calibration process. Since $\transform{robot}{ips}$ is therefore constant, prior to estimating $\transform{robot}{ips}$, we take $\frsub{P}{front}$ and $\frsub{P}{rear}$ as their respective averages over 16 consecutive IPS measurements.

We have calibrated $\Proj$ and used the IPS to estimate $\transform{robot}{ips}$ and $P_{\text{beacon}}$. The remaining unknowns in \eqref{eq_beacon_proj} are the desired transformation $\transform{cam}{robot}$ and $p_{\text{beacon}}$. $p_{\text{beacon}}$ is the beacon's observed position within the image, which we estimate by manually annotating the center of each beacon within the calibration images. $\transform{cam}{robot}$ may now be estimated with a PnP solver; we use OpenCV’s \verb|solvePnPRansac()| \cite{solvepnpransac}. Although \verb|solvePnPRansac()| accounts for lens distortion, to prevent rectification from propagating small errors in human image labels, we rectify the calibration images prior to annotation and pass null distortion coefficients to \verb|solvePnPRansac()|.

\subsubsection{Planar Constraint}
Since the IPS position estimates are only accurate to $\pm2$ cm in each direction \cite{marvel_mind}, their uncertainty dominates the error in this system. Therefore, we improve calibration accuracy by placing the observed beacons on two parallel planes (the floor and a table) and replacing the estimated $z$ coordinates of each group of beacons with their respective averages over all calibration samples. We find that this reduces the reprojection error of the estimated $\transform{cam}{robot}$ (see Section \ref{section:camera_ips_calibration}).

\subsection{Automatic Generation of Ground Truth}

\subsubsection{Image Labels}
We construct $\transform{obj}{ips}$ from two IPS beacons placed on each object of interest. However, this only offers object pose. Deriving full bounding boxes requires object dimensions; we manually measure these prior to data collection. We place the beacons on top of each object of interest and compute the center of the object in the IPS frame as:

\begin{equation}
\frsub{C}{obj} = \frac{1}{2}(\frsub{P}{front} + \frsub{P}{rear}) -
\begin{bmatrix}
0 & 0 & h_{\text{obj}} / 2
\end{bmatrix}\transpose
\end{equation}

where $h_{\text{obj}}$ denotes the height of the object. The global IPS coordinates of the bounding box vertices are computed straightforwardly as offsets from the object center. Finally, these vertices are transformed into the camera frame according to:

\begin{equation}
V_{\text{cam}} = \transform{cam}{robot} \cdot \transform{robot}{ips} \cdot V_{\text{ips}}
\end{equation}

where $V_{\text{ips}} \in \mathbb{R}^{4 \times 8}$ is a matrix of homogeneous object vertices. The image projections of the object vertices are computed as:

\begin{equation}
V_{\text{image}} = PV_{\text{cam}}
\end{equation}

We now convert $V_{\text{image}}$ to 2D Euclidean coordinates (such that $V_{\text{image}} \in \mathbb{R}^{2 \times 8}$) and compute the 2D bounding box label from the extrema of the projected 3D vertices:

\begin{equation}
(u_0, v_0) = (\min(\begin{bmatrix}1 & 0\end{bmatrix} \cdot V_{\text{image}}),
\min(\begin{bmatrix}0 & 1\end{bmatrix} \cdot V_{\text{image}}))
\end{equation}

\begin{equation}
(u_1, v_1) = (\max(\begin{bmatrix}1 & 0\end{bmatrix} \cdot V_{\text{image}}),
\max(\begin{bmatrix}0 & 1\end{bmatrix} \cdot V_{\text{image}}))
\end{equation}

where $(u_0, v_0)$ and $(u_1, v_1)$ denote the upper left and lower right corners of the bounding box label, respectively.

\subsubsection{Point Cloud Labels}

Point cloud bounding boxes are generated by transforming the 3D camera frame annotations into the LiDAR frame:

\begin{equation}
V_{\text{lidar}} = \transform{lidar}{cam} \cdot V_{\text{cam}}
\end{equation}

\subsection{Label Postprocessing}
\label{section:ransac_lidar_refinements}
\begin{figure}
\centering
\begin{tabular}{@{}p{0.45\linewidth}@{\quad}p{0.45\linewidth}@{}}
    \subfigimg[width=\linewidth]{\textcolor{white}{(a)}}{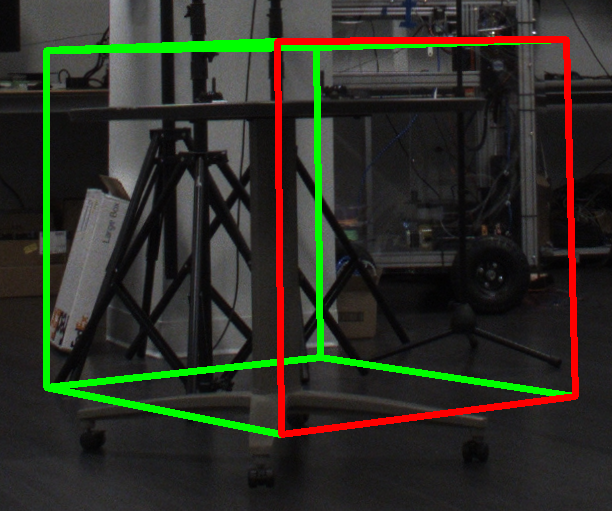} &
    \subfigimg[width=\linewidth]{\textcolor{white}{(b)}}{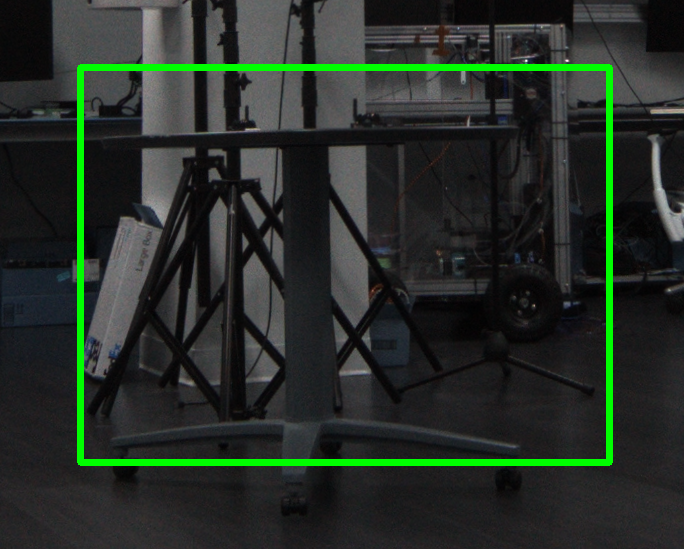} \\
  \end{tabular}

\caption{(a) Unrefined 3D bounding box label observed from camera frame. The front face is colored red. (b) Corresponding 2D box.}
\label{image_labels}
\end{figure}

\begin{figure}[h]
\includegraphics[width=\linewidth]{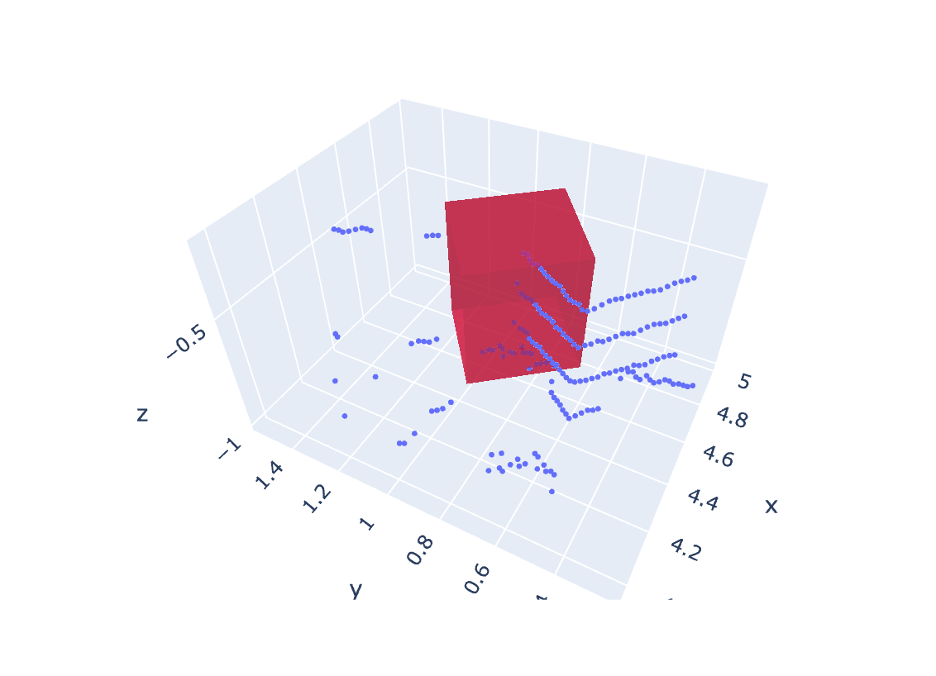}
\caption{Subset of point cloud shown to highlight the LiDAR label (red prism) prior to refinement. The object of interest depicted here is a filing cabinet.}
\label{lidar_label}
\end{figure}

We observe that generated LiDAR labels appear to have appreciably greater error than image labels (see Fig. \ref{image_labels}, \ref{lidar_label}). To improve the accuracy of generated LiDAR labels and reduce the pipeline’s reliance on the quality of LiDAR-camera calibration, we adapt RANSAC to refine LiDAR annotations.

Our refinement system leverages an extended RANSAC algorithm that samples from multiple model proposal function (MPF) definitions and uses a custom fitness function. MPF definitions, described in Section \ref{section:model-proposal-functions}, generate bounding boxes from sampled points. The fitness function (see Section \ref{section:fitness-function}) evaluates the quality of MPF-generated bounding boxes. The extended RANSAC algorithm (see Section \ref{section:ransac}) iteratively samples boxes and keeps the best box according to the fitness function.

We opt for RANSAC in lieu of ICP because we use a 16-channel LiDAR in our setup---our point clouds are too sparse for ICP to work reliably. Additionally, unlike ICP, RANSAC does not require a high-fidelity reference scan of the object of interest. Instead, we define MPFs to match the shape of each object of interest. Refining object labels reduces to designing MPFs for each object of interest.

\subsubsection{Model Proposal Functions}
\label{section:model-proposal-functions}

We design a set of tailored MPFs to refine the labels for each object class. For each iteration of RANSAC refinement, we randomly select a class-appropriate MPF and three points from within a distance $r$ of the center of the unrefined annotation. Points in the ground plane---estimated through RANSAC plane fitting---are removed to prevent the RANSAC refinement from preferring regions of dense ground points. Given the three sampled points, the MPF returns a plausible bounding box appropriate for the object of interest. The bottom face of the bounding box is constructed to lie in the ground plane. MPF bounding box derivation is further detailed in the appendix.

\subsubsection{Fitness Function}
\label{section:fitness-function}
Our fitness function is motivated by the observation that a good bounding box label to fit point cloud data contains many points near its faces. Therefore, our fitness is evaluated as follows:

Denote as $S$ all points within a shell of thickness $\pm\delta$ (measured along the normal of each of the box’s faces) surrounding the box. Define $A, B, C \subseteq S$ corresponding to the sets of points within the shell near each of the three different pairs of parallel faces. Note that this does \textit{not} define a partition of $S$; points near edges may be within distance $\delta$ of multiple faces. We compute the fitness score as:

\begin{equation}
f(S) = |A| + |B| + |C|
\end{equation}

\begin{figure}
    \centering
    \includegraphics[width=0.4\linewidth]{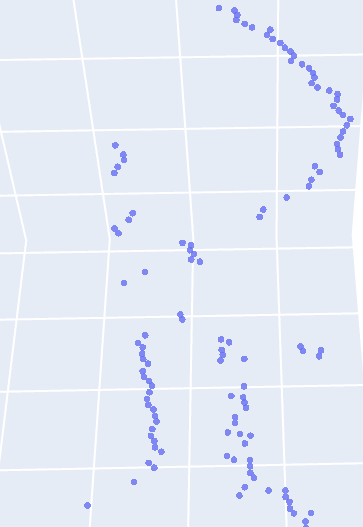}
    \caption{The surface of the table facing the LiDAR reflects beams back. However, the side of the table facing away from the LiDAR contains no points, as the LiDAR beams cannot reach it. Though the back side appears flat, it has low point density, so our fitness function does not favor bounding boxes along this edge.}
    \label{table_fig}
\end{figure}

This fitness function encourages bounding box proposals that align with objects’ true edges, which reflect many points, rather than object regions that are flat because an occlusion has blocked all points beyond a certain threshold. This is because these ``flat due to occlusion'' regions cut through objects (see Fig. \ref{table_fig}), and therefore, few points are sampled along these cutting planes. Our fitness function prioritizes placing bounding box faces closer to true object surfaces, which generally have denser LiDAR point coverage.

\subsubsection{Extended RANSAC}
\label{section:ransac}
We generalize RANSAC with a minor extension, allowing it to randomly select different model proposal functions. This allows RANSAC to fit objects seen from different angles. The extended algorithm is outlined below:

\begin{algorithm}[H]
\caption{Generalized RANSAC}
\begin{algorithmic}[1]
\Function{RANSAC}{\{$\mathbb{R}^3$\} $pcd$, \{MPF\} $funcs$, $n \in \mathbb{N}$}
    \State {$mpf \gets f \in funcs \text{, chosen at random}$}
    \State {$p \gets P \subseteq pcd \text{, chosen at random}$}
    \State {$best\_model \gets mpf(p)$}
    \State {$best\_fitness \gets fitness(best\_model)$}
    \For{$i \in \underline{n}$}
        \State {$mpf \gets f \in funcs \text{, chosen at random}$}
        \State {$p \gets P \subseteq pcd \text{, chosen at random}$}
        \State {$new\_model \gets mpf(p)$}
        \State {$new\_fitness \gets fitness(new\_model)$}
        \If {$new\_fitness > best\_fitness$}
            \State {$best\_fitness \gets new\_fitness$}
            \State {$best\_model \gets new\_model$}
        \EndIf
    \EndFor
    \State \Return $best\_model$
\EndFunction
\end{algorithmic}
\end{algorithm}

\section{Experiments}

\subsection{Camera-IPS Calibration}

To assess the effectiveness of imposing planar constraints in improving IPS-camera calibration, we compare calibration results with and without it (see Tab. \ref{calibration_table}). Additionally, we compare results with and without RANSAC \cite{source_camera_calibration}. For each method, we use SQPNP \cite{terzakislourakis2020} as the PnP solver and a virtual visual servoing (VVS) scheme \cite{chaumette2006} \cite{marchand7368948} for refinement. RANSAC is chosen for its robustness to outliers, while SQPNP serves as a baseline that fits all the correspondences. We use the OpenCV implementations of each algorithm. We evaluate RANSAC with two reprojection error thresholds for selecting inliers: $\delta=8$ and $\delta=25$ (in pixels).

\begin{table}[ht]
\begin{center}
\caption{\textsc{Quantitative Comparison of PnP Solvers}}
\begin{tabular}{|c|c|c|c|}
\hline

\textbf{Method} & Planar & \# Inliers & \textbf{RMSE}\\
& Constraint? & (out of 63) & \textbf{(px)}\\
\hline
RANSAC ($\delta = 8$) & no & 32 & \textbf{3.53282} \\
\hline
RANSAC ($\delta = 8$) & yes & 57 & \textbf{2.39437} \\
\hline
RANSAC ($\delta = 25$) & no & 55 & \textbf{7.51034} \\
\hline
RANSAC ($\delta = 25$) & yes & 63 & \textbf{3.23265} \\
\hline
SQPNP & no & - & \textbf{10.6648} \\
\hline
SQPNP & yes & - & \textbf{3.23265} \\
\hline
\end{tabular}
\label{calibration_table}
\caption*{Comparison between different PnP solvers for camera-IPS calibration. Results are also compared with and without our proposed planar constraint. For RANSAC, reported root-mean-square error (RMSE) values are only evaluated over points selected as inliers.}
\end{center}
\end{table}

\begin{figure}[ht]
    \centering
    \begin{tikzpicture}
    [spy using outlines={rectangle, magnification=6, size=0.3\linewidth, connect spies}]
    \node { \includegraphics[width=0.9\linewidth]{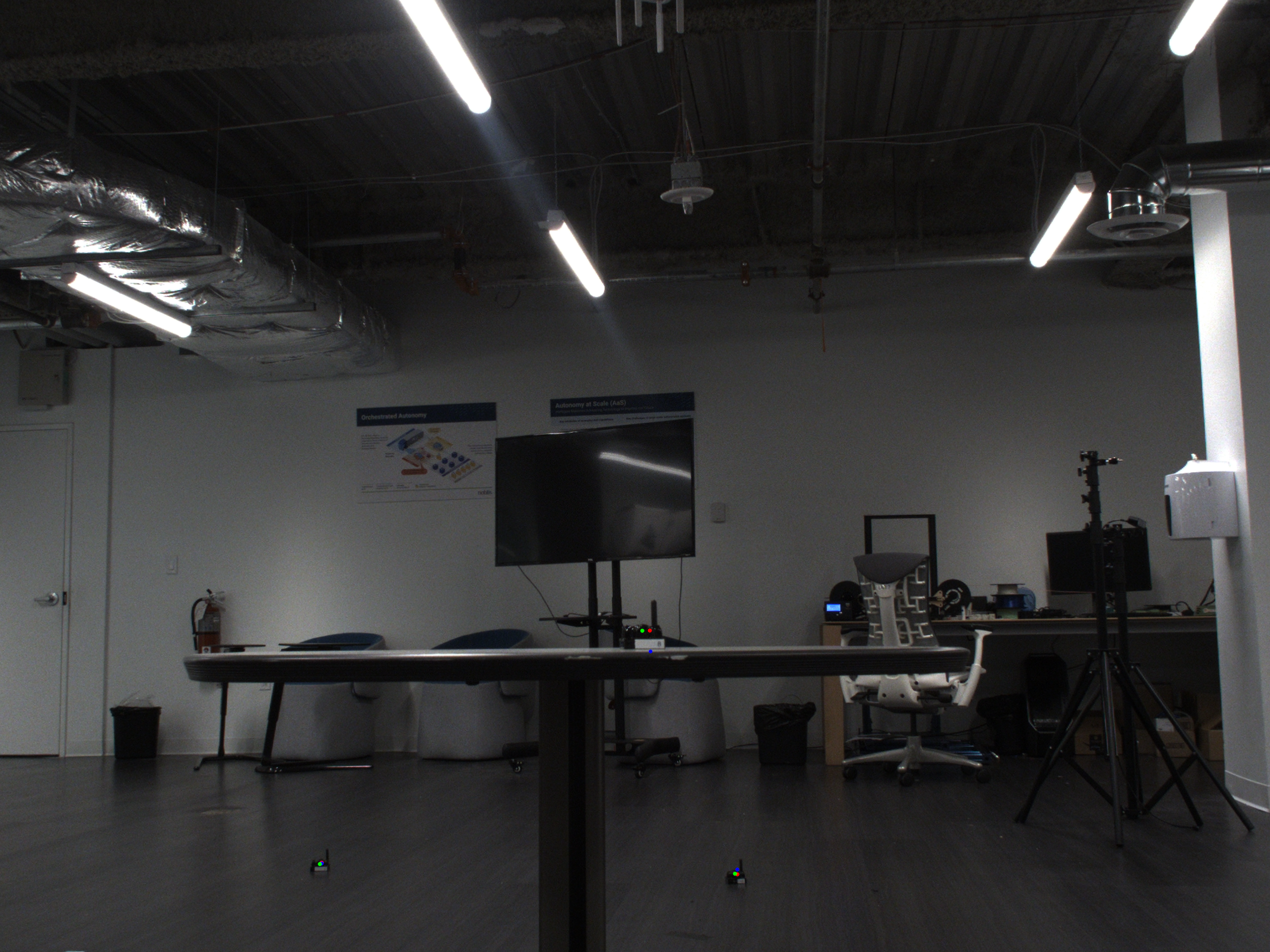} };
    \spy [red, magnification=11] on (-1.975, -2.42) in node at (-3, -4.6);
    \spy [red] on (0.08, -0.95) in node at (0, -4.6);
    \spy [red, magnification=11] on (0.63, -2.49) in node at (3, -4.6);
    \end{tikzpicture}
    \caption{Reprojected IPS points. Blue: reprojection after calibration (RANSAC, $\delta=8$) without planar constraint. Red: reprojection after calibration (RANSAC, $\delta=8$) with planar constraint. Green: manually labeled ground truth.}
\end{figure}

\subsection{Point Cloud Label Refinements}
To measure the effectiveness of our refinement strategy and its robustness to missing data, we select one of the cabinet training samples and down-sample the point cloud in the immediate region surrounding the unrefined annotation at various sampling proportions. We use RANSAC to refine the annotation based on the down-sampled point cloud. We conduct the experiment on 50 random subsamples of the point cloud for each sampling proportion, ensuring that our strategy remains effective regardless of which points are missing. For each trial, we record the fitness of RANSAC’s best proposal, evaluated on the original point cloud.

\begin{figure}[ht]
\centering
\includegraphics[width=\linewidth]{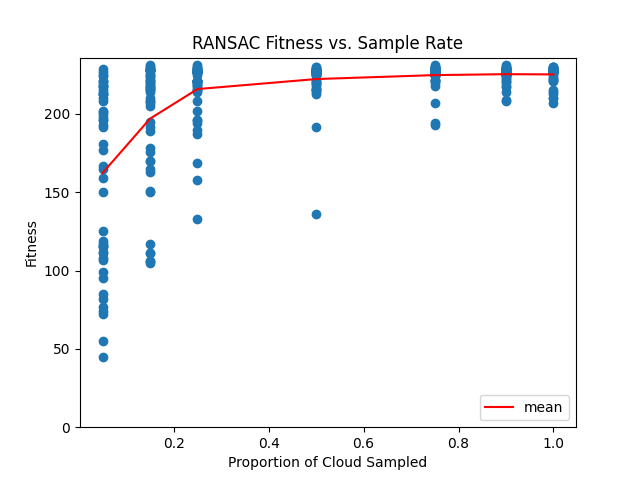}
\caption{Fitness function plotted as a function of sample rate. The blue dots represent individual trials, while the red line passes through the average fitness of each proportion.}
\label{RANSAC_fig}
\end{figure}

\begin{figure}[ht]
\centering
\begin{tabular}{@{}p{0.45\linewidth}@{\quad}p{0.45\linewidth}@{}}
    \subfigimg[width=\linewidth]{\textcolor{black}{(a)}}{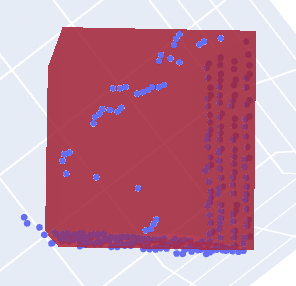} &
    \subfigimg[width=\linewidth]{\textcolor{black}{(b)}}{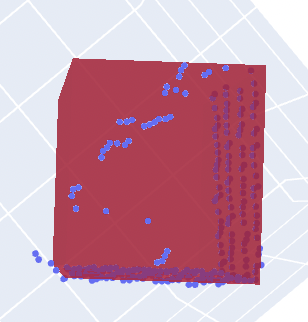} \\
  \end{tabular}
\caption{(a) Refined label estimated from 5\% of surrounding points. (b) Refined label estimated without down-sampling.}
\end{figure}

\begin{figure}[ht]
\centering
\begin{tabular}
{@{}p{0.45\linewidth}@{\quad}p{0.45\linewidth}@{}}
    \subfigimg[width=\linewidth]{\textcolor{white}{(a)}}{32_image_cropped.png} &
    \subfigimg[width=\linewidth]{}{32_lidar_larger_cropped.png} \\
    \subfigimg[width=\linewidth]{\textcolor{white}{(b)}}{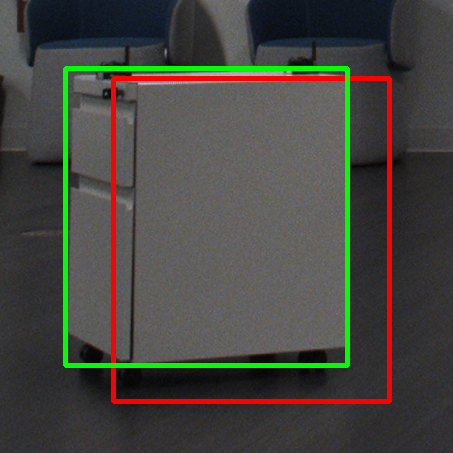} &
    \subfigimg[width=\linewidth]{}{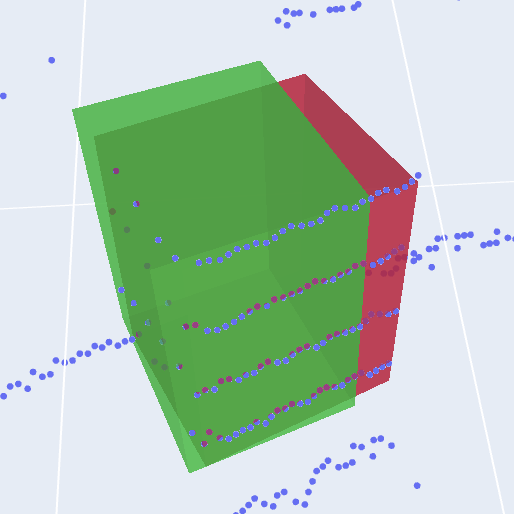} \\
    \subfigimg[width=\linewidth]{\textcolor{white}{(c)}}{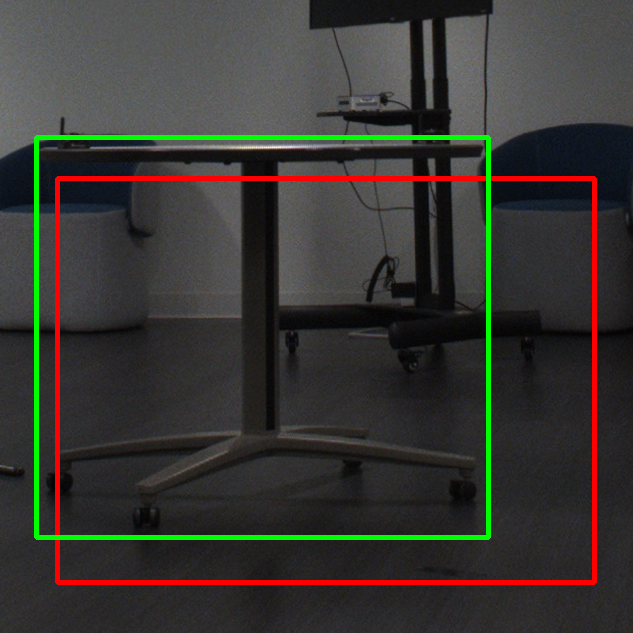} &
    \subfigimg[width=\linewidth]{}{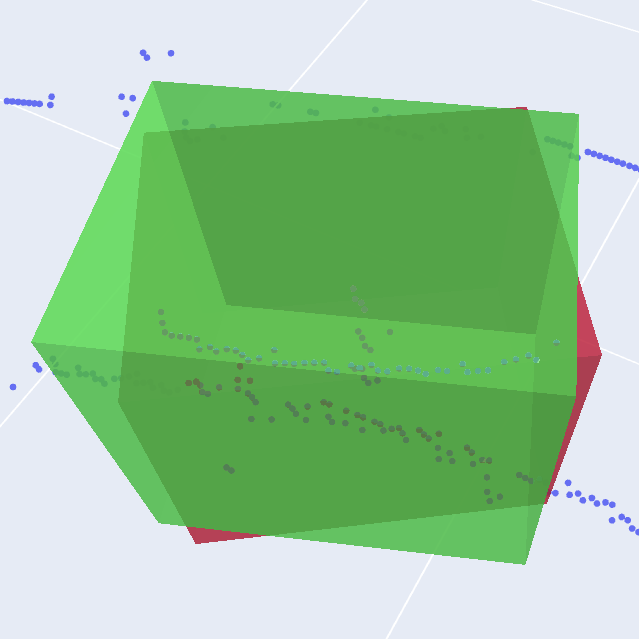} \\
    \subfigimg[width=\linewidth]{\textcolor{white}{(d)}}{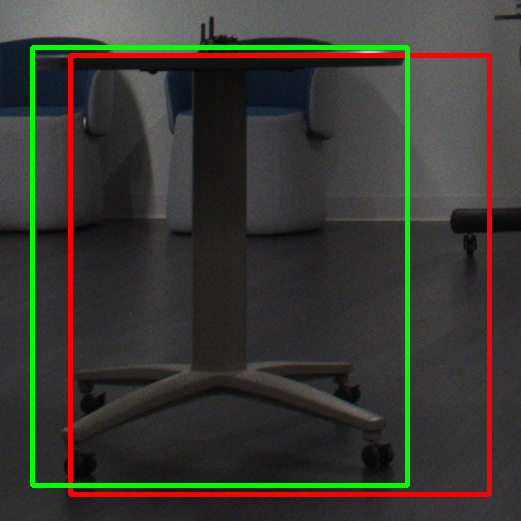} &
    \subfigimg[width=\linewidth]{}{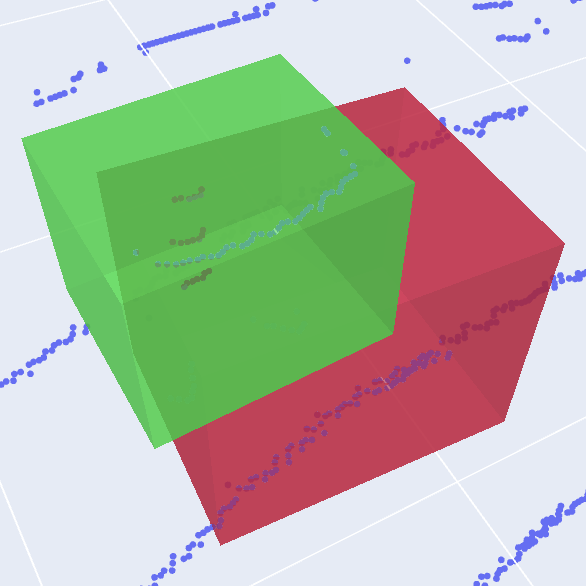} \\
  \end{tabular}
     
        \caption{Qualitative comparison between automated and manual labels (best viewed in color). Automated annotations are shown in red; manual annotations are green. (a), (b), and (c) show that the automated annotations often appear to align better with collected point clouds. (d) demonstrates the failure of the RANSAC refinement when a viewpoint is not covered by the model proposal functions.}

\label{qual_fig}
\end{figure}

\subsection{Annotation Speedup}

To quantify the reduction of manual effort, we collect 100 samples with the automated pipeline and compare the duration of manual and automatic annotation (see Tab. \ref{time_fig}). We use Label Studio \cite{label_studio} for image annotations and labelCloud \cite{sager2022labelCloud} for point cloud annotations. Sample collection time is dominated by time spent repositioning objects between samples. Since this process is arbitrary and difficult to control without introducing additional overhead and therefore biasing duration, we report the collection time via the automated pipeline for both. Our automated technique annotates the data 261.8 times faster than our human baseline, eliminates manual labeling effort, and lowers the total time spent creating the dataset by 61.5\%.

\begin{table}[ht]
    \begin{center}
    \caption{\textsc{Manual vs. Automated Duration (Minutes)}}
    \begin{tabular}{c|c|c|}
    \cline{2-3}
     & \multicolumn{2}{|c|}{\textbf{Step}}\\
     \hline
    \multicolumn{1}{|c|}{\textbf{Method}} & Collection & Post-Processing\\
    \hline
    \multicolumn{1}{|c|}{Manual} & 64.87 & 104.72\\
    \hline
    \multicolumn{1}{|c|}{Automated} & 64.87 & 0.4\\
    \hline
    \end{tabular}
    \caption*{Quantitative comparison of time to produce 100 samples. The post-processing step of our proposed method is automated and therefore requires no human effort.}
    \label{time_fig}
    \end{center}
\end{table}

\subsection{Annotation Quality}
With the data collected for the previous section, we use PyTorch3D's IoU function \verb|box3d_overlap()| \cite{ravi2020pytorch3d} to compute the average IoU between automated annotations and their manual counterparts. We find an average IoU of approximately $0.74$ for image annotations and $0.44$ for point cloud annotations. While an IoU of $0.44$ is admittedly low, our manual annotations are not perfect. We collect data with a 16-channel LiDAR, so our point clouds are particularly sparse. Furthermore, the table used in our experiment has a small LiDAR footprint, making manual annotation still more difficult. Lastly, the dimensions of manual annotations are not adjusted to match the measured ground truth. For these reasons, the manual annotations are prone to error (see Fig. \ref{qual_fig}).

\section{Discussion}

\subsection{Camera-IPS Calibration}

We combine RANSAC and SQPNP to reduce reprojection error arising from noise in the manually annotated calibration correspondences and the IPS readings. After imposing planar constraints, adjusting RANSAC’s inlier threshold to fit the last six correspondences increases the RMSE by over 33\%. This suggests that these six samples contribute substantially to the error and justifies the use of RANSAC. Additionally, in the constrained case, RANSAC ($\delta=25$) performs the same as the SQPNP baseline because RANSAC selects all the points as inliers.

\subsection{Quality of Unrefined Labels}
We attribute much of the label error to the calibration of $\transform{cam}{robot}$ and $\transform{lidar}{cam}$. Both methods employed here require human correspondence labeling, which introduces new sources of uncertainty. Additionally, we use a 16-channel LiDAR, making calibration more challenging.

We observed that the generated image labels are generally higher quality than their LiDAR counterparts prior to refinement. We believe there are two reasons for this.

First, the camera labels do not depend on $\transform{lidar}{cam}$, so it does not contribute to their error. In contrast, generating the LiDAR labels entails first transforming the robot-centric IPS measurements through $\transform{cam}{robot}$ and then through $\transform{lidar}{cam}$. Propagating positions through noisy rotations compounds small errors.

Second, the image labels are projected, unlike the point cloud labels. This means that an entire dimension of labeling error simply vanishes. Furthermore, perspective projection dictates that perturbing a 3D point affects its image by an amount scaling inversely proportionally with distance. This means that for farther objects, distance-independent sources of noise (such as beacon position estimates) contribute less to generated image labels than point cloud labels.




\section{Conclusion}
We have proposed an automated method to generate object detection annotations. To this end, we have presented a technique to calibrate a camera with an indoor positioning system and shown how imposing a geometric constraint improves calibration accuracy. Additionally, we have outlined an extension to RANSAC to refine point cloud bounding boxes even when large portions of objects of interest are missing. We believe that our approach will speed up data collection for future object detection research.

\section{Future Directions}
Future work should seek to propagate LiDAR refinements to the image labels and incorporate image information into the 2D annotations. Further, the LiDAR, camera, and IPS system should be jointly calibrated, shortening the transformation chain to the LiDAR frame. Lastly, using a 64-channel LiDAR would improve the robustness of point cloud annotation refinement.

\bibliography{sources} 


\section*{Appendix}
\label{section:appendix}
Our label postprocessing stage uses two different MPFs: an MPF designed for cabinets and an MPF designed for tables. They generate bounding boxes in slightly different ways.

\subsection{Cabinet MPF}
The ground plane is estimated from point cloud points within distance $r$ from the unrefined label. Denote the ground plane as $G$ and its unit normal $\hat{n}$. All model proposal functions first project the sampled points onto the ground plane, ensuring that the bottom face of the bounding box proposal is tangent to the ground. We call these projected points $P_1$, $P_2$, and $P_3$ and define unit vectors $\hat{v}_1$ and $\hat{v}_2$ as:

\begin{equation}
\hat{v}_1 = \frac{P_1 - P_3}{\vecnorm{P_1 - P_3}}
\end{equation}

\begin{equation}
\hat{v}_2 = \frac{P_2 - P_3}{\vecnorm{P_2 - P_3}}
\end{equation}

We aim to correct the angle between $\Vec{l}$ and $\Vec{w}$, forcing a right angle and ensuring a rectangular bounding box. To distribute the angular correction evenly, we construct $\Vec{l}$ and $\Vec{w}$ such that they are equal angles away from the angle bisector of $\hat{v}_1$ and $\hat{v}_2$. A vector bisector is constructed as follows:

\begin{equation}
\hat{s} = \frac{1}{\sqrt{2}}(\hat{v}_1 + \hat{v}_2)
\end{equation}

Because the sample points were projected onto $G$, $\hat{v}_1$ and $\hat{v}_2$ lie in $G$, and so does $\hat{s}$. Therefore, $\hat{s}$ is orthogonal to $\hat{n}$, so we may construct an orthogonal vector as:

\begin{equation}
\hat{o} = \hat{n} \times \hat{s}
\end{equation}

By constructing $\Vec{l}$ and $\Vec{w}$ to lie in $G$ and bisect the angles formed between $\hat{s}$ and $\hat{o}$ and $\hat{s}$ and $-\hat{o}$, respectively, we ensure that $\Vec{l}$ and $\Vec{w}$ are orthogonal both to each other and to $\hat{n}$. Therefore, we have:

\begin{equation}
\Vec{w} = \frac{w}{\sqrt{2}}(\hat{s} - \hat{o})
\end{equation}

\begin{equation}
\Vec{l} = \frac{l}{\sqrt{2}}(\hat{s} + \hat{o})
\end{equation}

where $w$ denotes the object's measured width and $l$ denotes the object's measured length.

\begin{figure}[h]
\includegraphics[width=\linewidth]{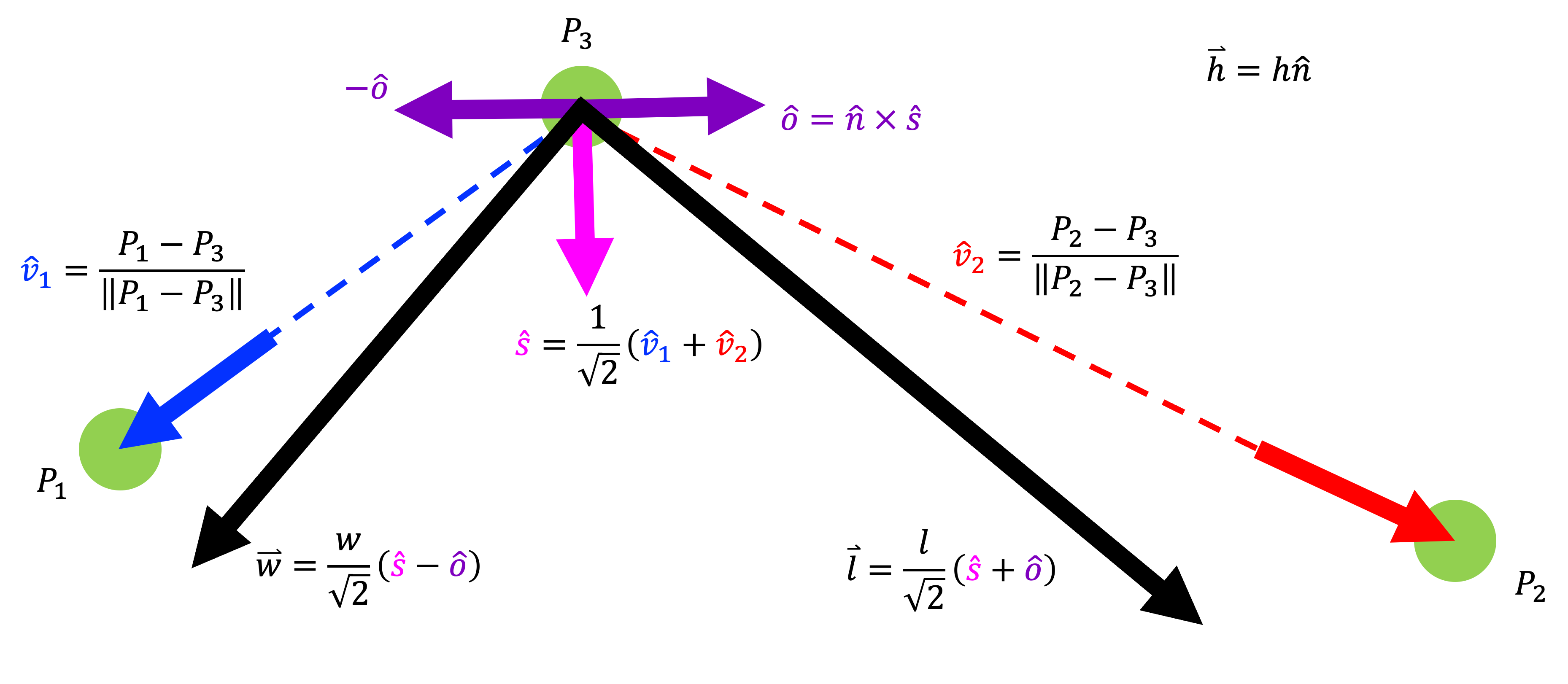}
\caption{Construction of orthogonal vectors from sample points.}
\end{figure}

Because the bounding box proposal must be tangent to the ground plane, we construct $\Vec{h}$ to be parallel to $\hat{n}$:

\begin{equation}
\Vec{h} = h_{\text{obj}}\hat{n}
\end{equation}


Finally, we construct the bounding box based on the assumption that $P_3$ is the left front vertex of the box. This is an example of an MPF-specific assumption. A similar MPF may be defined by assuming that $P_3$ is the \textit{right} front vertex of the box, allowing refinement across different object viewing angles. Furthermore, though we omit the details, the bounding box may also be modeled from a sample of two points along one face to account for instances where only one side of the box is facing the LiDAR.

\subsection{Table MPF}
The irregular shape and small LiDAR-facing surface area of the table we capture mean that the table requires its own assumptions and MPFs. We construct the bounding box as with the cabinet, except that we assume a point from the stem of the table, rather than a corner. Additionally, prior to bounding box refinement, we remove all points from the surrounding region that are below a minimum height $\delta$. Our goal is a bounding box that aligns with the table's edge. Removing unnecessary points such as the table legs and the ground plane improves the chance of sampling points from the rim of the table.

\end{document}